\documentclass[conference]{IEEEtran}
\IEEEoverridecommandlockouts
\usepackage{cite}
\usepackage{amsmath,amssymb,amsfonts}
\usepackage{algorithmic}
\usepackage{graphicx}
\usepackage{textcomp}
\usepackage{xcolor}

\usepackage{adjustbox}
\usepackage{fontawesome}
\usepackage[most]{tcolorbox}
\usepackage{tabularx}
\usepackage{hyperref}
\usepackage{subcaption}
\usepackage[T2A, T1]{fontenc}
\usepackage[utf8]{inputenc}
\usepackage[english]{babel}

\def\BibTeX{{\rm B\kern-.05em{\sc i\kern-.025em b}\kern-.08em
    T\kern-.1667em\lower.7ex\hbox{E}\kern-.125emX}}
\begin{document}

% ID Card command (original with user icon)
\newcommand{\IDCard}[7]{%
  {\footnotesize                      % <<< makes all text inside smaller
  \begin{tcolorbox}[
      colback=#7!10,
      colframe=black,
      width=\linewidth,
      arc=6pt,
      boxsep=5pt,
      left=6pt,right=6pt,top=6pt,bottom=6pt,
  ]
    \noindent
    % ---------- icon + basic section -----------------------
    \begin{minipage}[t]{1.5cm}          % icon column (fixed)
      \centering
      \raisebox{-15pt}{\fontsize{50}{50}\selectfont\color{black}\faUser}
    \end{minipage}%
    \hspace{1em}%
    \begin{minipage}[t]{\dimexpr\linewidth-1.8cm-1em\relax}% basic section
      \renewcommand{\arraystretch}{1.1}
      \begin{tabularx}{\linewidth}{@{}p{1.3cm}X@{}}
        Name:            & #1\\
        Age:             & #2\\
        Sex:             & #3\\
        Nationality:     & #4\\
        Group: & #5
      \end{tabularx}
    \end{minipage}

    % ---------- Outlook paragraph -------------------------
    \vspace{0.8em}
    \textbf{Outlook:} #6
  \end{tcolorbox}%
  }% end \small
}

% Prompt Card command
\newcommand{\PromptCard}[2]{%
  {\footnotesize
  \begin{tcolorbox}[
      colback=gray!15,
      colframe=black,
      width=\linewidth,
      arc=6pt,
      boxsep=5pt,
      left=6pt,right=6pt,top=6pt,bottom=6pt,
  ]
    
    % ---------- System prompt section (full width) -------------
    \vspace{1em}
    \textbf{System prompt:}
    
    \vspace{0.5em}
    #1
    
    % ---------- User prompt section (full width) -------------
    \vspace{1em}
    \textbf{User prompt:}
    
    \vspace{0.5em}
    #2
  \end{tcolorbox}%
  }% end \footnotesize
}

\title{Language, Culture, and Ideology: Personalizing \\Offensiveness Detection in Political Tweets with Reasoning LLMs
}

\author{
Dzmitry Pihulski and Jan Kocoń \\
\textit{Department of Artificial Intelligence, Wroclaw Tech, Poland} \\
\texttt{\{dzmitry.pihulski, jan.kocon\}@pwr.edu.pl}
}

% \author{
% \IEEEauthorblockN{1\textsuperscript{st} Dzmitry Pihulski}
% \IEEEauthorblockA{\textit{Department of Artificial Intelligence} \\
% \textit{Wroclaw University of Science and Technology}\\
% Wroclaw, Poland \\
% dzmitry.pihulski@pwr.edu.pl}
% \and
% \IEEEauthorblockN{2\textsuperscript{nd} Jan Kocoń}
% \IEEEauthorblockA{\textit{Department of Artificial Intelligence} \\
% \textit{Wroclaw University of Science and Technology}\\
% Wroclaw, Poland \\
% jan.kocon@pwr.edu.pl} 
% \and
% \and
% \and
% \and

\maketitle

\begin{abstract}
We explore how large language models (LLMs) assess offensiveness in political discourse when prompted to adopt specific political and cultural perspectives. Using a multilingual subset of the MD-Agreement dataset centered on tweets from the 2020 US elections, we evaluate several recent LLMs -- including DeepSeek-R1, o4-mini, GPT-4.1-mini, Qwen3, Gemma, and Mistral -- tasked with judging tweets as offensive or non-offensive from the viewpoints of varied political personas (far-right, conservative, centrist, progressive) across English, Polish, and Russian contexts. Our results show that larger models with explicit reasoning abilities (e.g., DeepSeek-R1, o4-mini) are more consistent and sensitive to ideological and cultural variation, while smaller models often fail to capture subtle distinctions. We find that reasoning capabilities significantly improve both the personalization and interpretability of offensiveness judgments, suggesting that such mechanisms are key to adapting LLMs for nuanced sociopolitical text classification across languages and ideologies.
\end{abstract}


\section{Introduction}
Detecting offensive language is vital for fostering respectful discourse, particularly on social media. Yet, offensiveness is inherently subjective -- shaped by individual ideologies, cultural backgrounds, and values \cite{sap2019risk, leonardelli-etal-2021-agreeing}. Supervised models rely on ground-truth labels that reflect annotators' biases, complicating efforts to build fair and robust systems \cite{binns2020apparent}. Political discourse, marked by polarization, offers a compelling lens: individuals often tolerate combative language from their own side while labeling opposing views as offensive \cite{bail2018exposure}.

Recent advances in large language models (LLMs) -- from GPT-based to instruction-tuned variants -- have improved context-sensitive understanding \cite{vaswani2017attention, devlin2019bert, brown2020language, grattafiori2024llama}. But standard LLMs often align with general consensus, limiting their ability to reflect individual interpretations of offensive speech. Personalization demands not only semantic comprehension but also simulation of reasoning grounded in specific ideological and cultural viewpoints \cite{nadeem2021stereoset, davani2024disentangling}. Personalization in NLP is a well-studied challenge for classical models, including encoder-only transformers \cite{kocon2021offensive,kocon2021learning,kazienko2023human,milkowski2021personal, kanclerz2021controversy, kanclerz2022if, ngo2022studemo, milkowski2022multitask, mieleszczenko2023capturing, kanclerz2023pals, kanclerz2023towards, milkowski2023modeling, kocon2023differential, kocon2023multi, ngo2025integrating, ferdinan2025fortifying}. However, research on personalization for large language models (LLMs) remains limited and underexplored \cite{kocon2023chatgpt,wozniak2024personalized,wozniak2025improving}.

This work explores whether reasoning helps LLMs better personalize offensiveness detection, particularly in politically charged contexts. We hypothesize that reasoning-equipped models more effectively simulate perspective-taking when prompted with ideological and cultural cues. We compare SOTA models on a corpus of political tweets translated into English, Polish, and Russian. These include large reasoning-enabled models (DeepSeek-R1, OpenAI o4-mini), large non-reasoning model (DeepSeek-V3), and smaller ones (Qwen3-8B with reasoning, Qwen3-4B, Gemma 3-4B-IT, Mistral 7B Instruct v0.3, GPT-4.1-mini).

We build detailed ideological profiles (far-right, moderate conservative, progressive left, centrist) to assess how simulated political alignment and nationality affect model judgments. Our evaluation highlights both the promise and pitfalls of incorporating explicit reasoning into LLMs for personalized classification of offensive language.

\noindent Our main contributions are:
\begin{itemize}
\item We introduce a novel framework for evaluating personalization in offensiveness detection across political, ideological, and cultural dimensions.
\item We demonstrate that reasoning mechanisms significantly enhance LLMs' ability to simulate diverse ideological perspectives.
\item We provide a comparative analysis of state-of-the-art models, showing the superior performance of reasoning-enabled models like DeepSeek-R1 and OpenAI’s o4-mini.
\item We release a multilingual dataset derived from the MD-Agreement corpus for reasoning and personalization in offensive language classification.
\end{itemize}

\section{Related Work}

Offensive language detection has drawn attention for its role in curbing online harm and promoting civil discourse \cite{schmidt2017survey, davidson2017automated, li2024hot, zampieri2023offenseval, Ge2025}. Early work relied on supervised learning with annotated datasets, though these often encoded annotators' biases and judgments \cite{waseem-hovy-2016-hateful, waseem2016you, davidson2017automated}. Later research highlighted the inadequacy of general models in capturing individual and contextual nuances in perceived offensiveness \cite{sap2019risk, kennedy2020contextualizing, binns2020apparent}.

Transformer-based models like BERT and GPT have significantly advanced text classification tasks, including offensive speech detection \cite{vaswani2017attention, devlin2019bert, brown2020language, achiam2023gpt, vidgen2020directions, caselli2021hatebert}. However, their training on broad corpora limits their ability to model user-specific, personalized, or culturally grounded interpretations of offensive content \cite{sap2020social, leonardelli-etal-2021-agreeing, kocon2021offensive}.

Incorporating cultural and ideological variation has proven essential. The MD-Agreement dataset \cite{leonardelli-etal-2021-agreeing} revealed annotator disagreement rooted in ideological diversity, while personalization improves performance over general models \cite{kocon2021offensive}.

Recent work also explores the role of explicit reasoning in LLMs \cite{wiegreffe2021teach, tian2024role}. Models that generate reasoning traces improve both performance and interpretability in tasks involving bias and stereotypes \cite{nadeem2021stereoset, davani2024disentangling}, with intermediate reasoning steps shown to enhance classification consistency \cite{wang2022self, tian2024role}.

Cross-cultural perspectives are increasingly emphasized in NLP, recognizing that perceptions of offensiveness vary widely across languages and societies \cite{ranasinghe-zampieri-2021-mudes, mnassri2024survey}. Multilingual models like XLM-R \cite{conneau2020unsupervised} have enabled more culturally sensitive and nuanced analyses \cite{shi2022cross, ranasinghe2021multilingual}.

Our work builds on these developments by examining how reasoning-enabled LLMs handle political ideology, cultural context, and multilingualism in personalized offensiveness detection. We aim to address persistent gaps by explicitly modeling these factors in both input prompts and evaluation.

\section{Data}
\subsection{Source Dataset}
For our experiments, we used a subset of the MD-Agreement dataset \cite{leonardelli-etal-2021-agreeing}. The MD-Agreement dataset contains over 10,000 tweet IDs, each labeled as offensive or non-offensive by five different annotators via Amazon Mechanical Turk. The tweets are drawn from multiple domains (MD), including COVID-19, BLM (Black Lives Matter), and the US 2020 elections. The dataset includes the tweet text, individual annotations, and anonymized annotator IDs. For our research, we used a subset of 300 samples from the US 2020 elections domain, without relying on the provided annotations.

\subsection{Translations}
Each tweet in our subset was translated from the original language (English) into Polish and Russian using the DeepL API \cite{deepl_api_2025}. Not every translation was accurate. After manual review, we found that approximately 7\% of translations contained errors, most commonly where only the first part or sentence was translated, while the rest remained in the original language. Since tweets on X (Twitter) can contain user tags, all user references in the dataset were replaced with the placeholder “\texttt{<user>}” to protect personal information. However, these placeholders sometimes confused the DeepL API, resulting in untranslated tweets when “\texttt{<user>}” tags appeared at the start of the text. After manually correcting the translations, we reduced the error rate to just 1\% (three samples). Tweets with ambiguous meanings or unclear abbreviations were left untranslated and were not used in further research. Examples include: “\texttt{<user>} \texttt{<user>} Deez Nuts,” “\texttt{<user>} \texttt{<user>} AIMS???U+1F974” and “\texttt{<user>} \texttt{<user>} \texttt{<user>} Just take the L g, or find better evidence”.

\section{Experimental Setup}
\subsection{Language Models and Inference}
To evaluate the influence of model size and reasoning capabilities on classification behavior, we compared four categories of language models based on two variables: model size (small vs. large) and reasoning capability (enabled vs. disabled). 

\begin{itemize}
  \item \textbf{Large models with reasoning:} DeepSeek-R1~\cite{guo2025deepseek} and OpenAI's o4-mini -- o4-mini-2025-04-16~\cite{openai2025o4mini}. While o4-mini is a closed-source model with unspecified size, DeepSeek-R1 has over 600B parameters~\cite{guo2025deepseek}.
  \item \textbf{Large model without reasoning:} DeepSeek-V3~\cite{liu2024deepseek}.
  \item \textbf{Small models with reasoning:} Qwen3-8B, evaluated with reasoning capabilities enabled~\cite{qwen3}.
  \item \textbf{Small models without reasoning:} Qwen3-4B (with reasoning capabilities explicitly disabled), Gemma 3 4B IT, Mistral 7B Instruct v0.3 and GPT-4.1-mini -- gpt-4.1-mini-2025-04-14~\cite{qwen3, gemma_2025, mistral7b, openai2025gpt41}.
\end{itemize}

This categorization allowed us to examine the interplay between reasoning ability and model capacity in handling offensive content classification across different political and linguistic contexts.

% Each model was asked to classify a tweet into one of two categories -- \textit{offensive} or \textit{not offensive} -- as it might be judged by a person described in the prompt. The prompt, including detailed personality descriptions, is provided in the Appendix (see Fig.~\ref{fig:all_prompts}, \ref{fig:12personalities}).

Each model was asked to classify a tweet into one of two categories -- \textit{offensive} or \textit{not offensive} -- as it might be judged by a person described in the prompt. The prompt, including detailed personality descriptions, is provided in the GitHub repository (see the \texttt{data/personalities/}, \texttt{data/prompts/} in the repository~\cite{repo2025}).

Another variable modified in the prompts was the nationality of the simulated individual. To explore how cultural background may influence classification, we varied both the language and the stated nationality of the person in the prompt. Specifically, American nationality was paired with English prompts, Polish nationality with Polish translations, and Russian nationality with Russian translations. This allowed us to examine whether the model's judgments differed when simulating individuals from distinct national contexts.

\begin{figure}
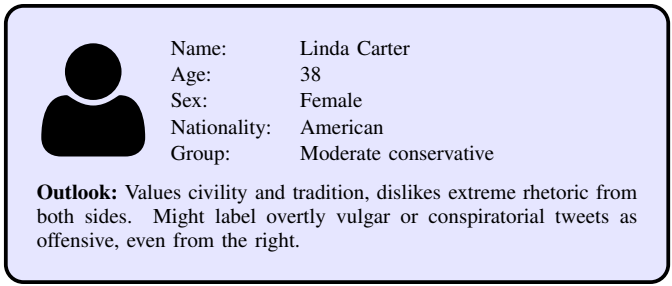

    \centering
  \begin{adjustbox}{width=\columnwidth}
    \IDCard{Linda Carter}
           {38}
           {Female}
           {American}
           {Moderate conservative}
           {Values civility and tradition, dislikes extreme rhetoric from both sides. \
            Might label overtly vulgar or conspiratorial tweets as offensive, even from the right.}{blue}
  \end{adjustbox}
    \caption{Example personality representation used in the prompt. The model receives not only the name, age, sex, and political affiliation, but also the nationality, allowing us to explore how individuals from different nationalities might respond to the same text.}
    \label{fig:Personality}
\end{figure}

To explore different perspectives, four personality profiles representing diverse political groups were created: far-right conservative, moderate conservative, progressive left, and centrist/independent. In addition to political affiliation, each profile included other personality traits. As a result, the large language models were given a well-rounded representation of each hypothetical person (see Fig.~\ref{fig:Personality}).

The answers from the models with reasoning capabilities were collected five times for each sample to reduce the impact of randomness in the response distribution. Ideally, this randomness could be eliminated by setting the temperature parameter to \( 0 \); however, this option was not available for these models. In contrast, the answers from smaller models without reasoning capabilities were collected in the form of probabilities for predicting $0$ or $1$. The final classification was determined based on which outcome had the higher probability.

All datasets and models used in this study are publicly available under open licenses for research purposes. However, the datasets may include harmful content such as offensive language or personal information. Its presence underscores the importance of ethical filtering and responsible handling when working with large-scale web data for training language models.

\subsection{Data Preprocessing and Analysis}
For our experiments, we used an H200 GPU for 7 hours and an H100 GPU for 10 hours, both accessed via the \href{https://vast.ai}{Vast.ai} platform. Implementation and analysis were conducted using Python libraries including transformers, pandas, matplotlib, scipy, numpy, scikit-learn, and statsmodels. The experiment was structured so the model's answer is binary: $1$ for an offensive tweet and $0$ for a non-offensive one. For simplicity, outputs (collected five times for reasoning-enabled models) are treated as samples from a binomial distribution, with each answer a Bernoulli trial $B(1, p)$. Formally, the model's response for each input prompt $n_i$ is given by:

\begin{equation*}
\adjustbox{max width=0.93\columnwidth}{$
  F(F_k(\dots F_2(F_1(n_i)) \dots)) =
    \begin{cases}
    1, & \text{with probability } p \\
    0, & \text{with probability } 1 - p
    \end{cases}
    $}
\end{equation*}
where:
\begin{itemize}
\item $p \in [0, 1]$ is the probability of generating the token that represents $1$ (offensive).
\item $F(x)$ denotes the overall function representing the Bernoulli trial.
\item $F_i(x),\ i \in \{1, \dots, k\}$, represents intermediate functions corresponding to multinomial distribution trials when sampling the next token in the model's reasoning process.
\item $n_i, i \in \{1, 3564\}$ is the order number of the prompt (297 tweets in 3 languages with 4 versions of personalities).
\end{itemize}

The final answer is not always a Bernoulli trial from the same binomial distribution. The probability $p$ is heavily influenced by the model's reasoning—e.g., if it deems a tweet offensive, $p$ increases. Since LLM reasoning can be seen as a sequence of multinomial trials, each answer is a sample from one of many underlying Bernoulli distributions. We formalize this idea and show the resulting distribution remains binomial under certain conditions.

Let $X$ be the outcome of a single composite trial where a model first selects an index $i \in \{1, \dots, n\}$ with probability $w_i$, and then draws $X \sim \text{Bernoulli}(p_i)$. By the law of total expectation, the overall probability of success is:

\begin{equation*}
\begin{split}
\mathbb{P}(X = 1) &= \sum_{i=1}^n w_i \cdot \mathbb{P}(X = 1 \mid i) =\\
&= \sum_{i=1}^n w_i p_i = p.
\end{split}
\end{equation*}

Thus, $X \sim \text{Bernoulli}(p)$, where $p$ is the weighted average of the individual success probabilities $p_i$. Repeating this process independently $m$ times yields a binomial distribution $\text{Binomial}(m, p)$.
To draw further conclusions, we need to estimate the true probability that the model assigns one of the two possible answers: an offensive or a non-offensive tweet. To do this, we generated answers to the same question five times and then estimated the probability $p_{n_k}$ by dividing the sum of positive outcomes by the count:

\begin{equation*}
\hat{p}_{n_k} = \frac{\sum_{i=1}^{5} X_i^{n_k}}{5}
\end{equation*}
where $X_i^{n_k}$ is the $i$-th answer for prompt $n_k$. After estimating $\hat{p}_{n_k}$, we calculated the Wald confidence interval (CI) at significance level $\alpha = 10\%$ as follows.
Given five independent observations from a binomial distribution Wald confidence interval is defined as follows:

\begin{equation*}
\adjustbox{max width=0.95\columnwidth}{$
    p_{n_k} \in \left[ \hat{p}_{n_k} - z_{1-\alpha/2} \sqrt{\frac{\hat{p}_{n_k}(1 - \hat{p}_{n_k})}{5}},\; \hat{p}_{n_k} + z_{1-\alpha/2} \sqrt{\frac{\hat{p}_{n_k}(1 - \hat{p}_{n_k})}{5}} \right]
    $}
\end{equation*}
where $z_{1-\alpha/2}$ is the is the $1-\frac{\alpha}{2}$ quantile of a standard normal distribution (e.g., $1.645$ for $90\%$ confidence).
In our scenario, where each response consists of 5 Bernoulli trials, the possible values of $\hat{p}_{n_k}$ and their corresponding Wald CIs at a significance level of $\alpha = 10\%$ are:

$$
\hat{p}_{n_k} =
\begin{cases}
0.0, & \text{Wald CI: } [0, 0] \\
0.2, & \text{Wald CI: } [0, 0.49] \\
0.4, & \text{Wald CI: } [0.03, 0.76] \\
0.6, & \text{Wald CI: } [0.23, 0.96] \\
0.8, & \text{Wald CI: } [0.51, 1.0] \\
1.0, & \text{Wald CI: } [1.0, 1.0] \\
\end{cases}
$$

In cases where $\hat{p}_{n_k} \in \{0.4, 0.6\}$, the Wald confidence intervals are too wide to confidently determine whether $p_{n_k} > 0.5$ or $p_{n_k} < 0.5$ at the given level of significance. As a result, these intervals provide insufficient evidence to infer the model's preference with respect to offensive classification.

\begin{figure*}[ht]
  \includegraphics[width=\columnwidth]{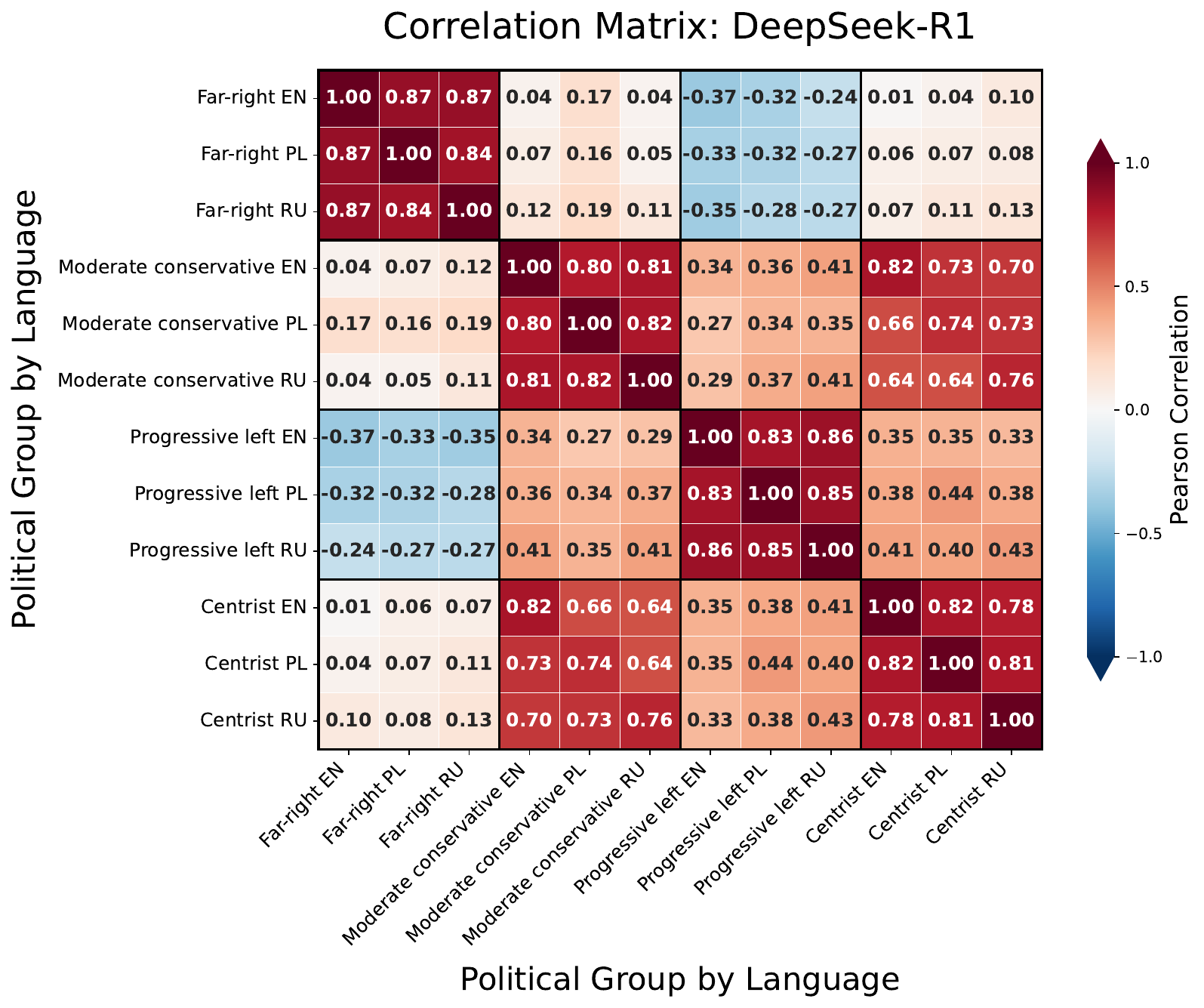} \hfill
  \includegraphics[width=\columnwidth]{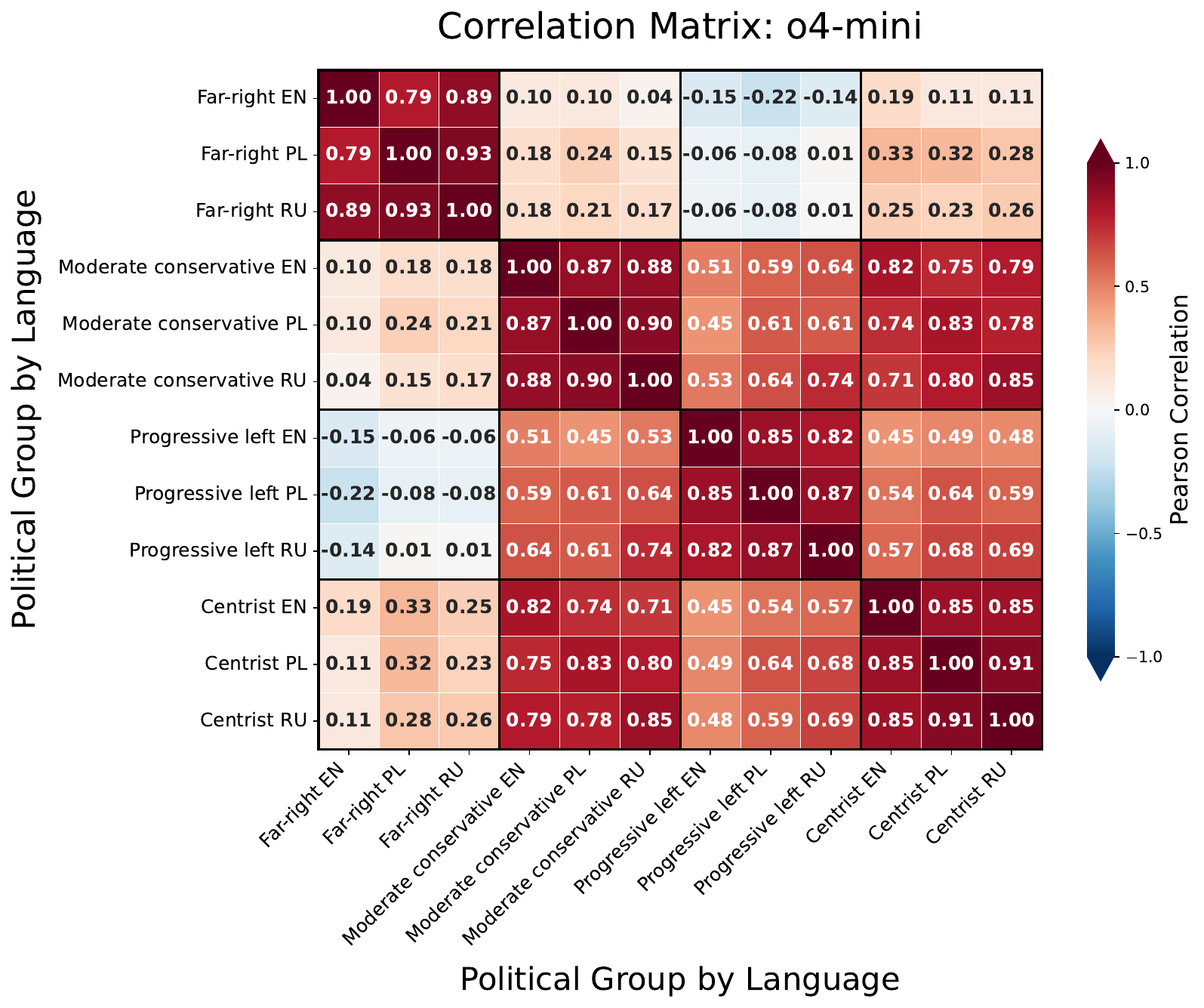}
  \caption{Correlation plots showing the agreement in tweet classifications across different political perspectives for two reasoning-enabled language models: DeepSeek-R1 and OpenAI's o4-mini. Each plot visualizes how consistently each model assigns offensive/non-offensive labels based on the provided personality profiles, highlighting alignment or divergence across political viewpoints.}
  \label{fig:reasoning_models}
\end{figure*}

After data collection, we excluded uncertain results using this rule: if for a sample \( n_k \), the estimated probability \( \hat{p}_{n_k} \in \{0.4, 0.6\} \), it was removed due to low classification confidence. The proportion of excluded samples was:

\begin{itemize}
    \item \textbf{DeepSeek-R1}: \( 9.3\% \) (331 samples excluded)
    \item \textbf{OpenAI's o4-mini}: \( 9.3\% \) (329 samples excluded)
    \item \textbf{Qwen3-8B}: \( 22.6\% \) (802 samples excluded)
    
\end{itemize}

The final step involved aggregating the statistically sufficient data into a unified dataset based on the following rule:  
\begin{itemize}
    \item Samples with \( \hat{p}_{n_k} \in \{0.0, 0.2\} \) were assigned a ground truth label of \( 0 \) (not offensive).
    \item Samples with \( \hat{p}_{n_k} \in \{0.8, 1.0\} \) were assigned a ground truth label of \( 1 \) (offensive).
\end{itemize}

\section{Results}

\subsection{Large Models with Reasoning Capabilities}
\label{bigger_reasoning_models}

% \begin{figure*}[ht]
%   \includegraphics[width=\columnwidth]{images/DeepSeek-R1/correlation_plot_deepseek_r1.pdf} \hfill
%   \includegraphics[width=\columnwidth]{images/o4-mini/correlation_plot_openai.pdf}
%   \caption{Correlation plots showing the agreement in tweet classifications across different political perspectives for two reasoning-enabled language models: DeepSeek-R1 and OpenAI's o4-mini. Each plot visualizes how consistently each model assigns offensive/non-offensive labels based on the provided personality profiles, highlighting alignment or divergence across political viewpoints.}
%   \label{fig:reasoning_models}
% \end{figure*}

We observe consistently high correlations ($r > 0.8$) among reasoning models (DeepSeek-R1, o4-mini) within the same political groups (main diagonal blocks), suggesting that language and nationality have minimal effect on model outputs. This stability aligns with the uniformity of the prompt structure across languages, where nationality was the only major variable.

Correlations within political groups remain stable across languages (EN, PL, RU), suggesting model judgments are consistent for a given ideology regardless of language. This is evident in the correlation matrices, where similar-intensity blocks highlight language-invariant ideological patterns.

Though not perfect, the observed correlations show strong cross-linguistic consistency. The far-right group has the highest internal coherence ($0.79 \leq r \leq 0.93$) and lowest cross-group correlations, especially with the progressive left ($-0.37 \leq r \leq 0.01$). Moderate conservatives show strong internal consistency ($0.80 \leq r \leq 0.90$) and align closely with centrists ($0.64 \leq r \leq 0.85$). The progressive left also exhibits high internal cohesion ($0.82 \leq r \leq 0.87$).

The far-right stands out with markedly lower intergroup correlations to all other groups ($-0.37 \leq r \leq 0.33$), reflecting distinct classification patterns and reinforcing the ideological divide—especially with the progressive left, where the most negative correlation was observed.

\begin{figure}[ht]
  \includegraphics[width=\columnwidth]{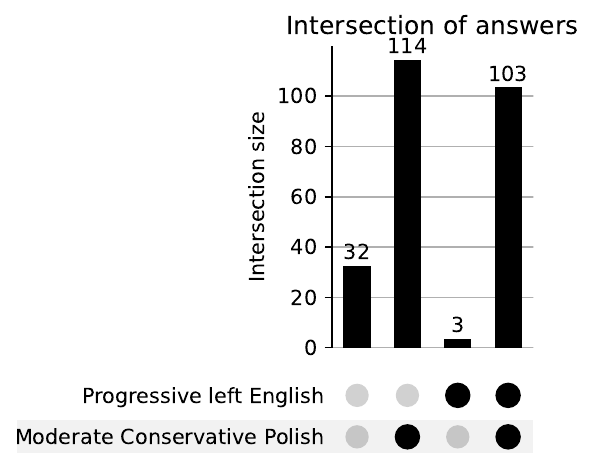}
  \caption{Upset plot showing the overlap in classification decisions between the 'Moderate Conservative Polish' and 'Progressive Left English' personalities using the DeepSeek-R1 model. Despite a relatively low Pearson correlation ($r = 0.27$), the two groups agree on approximately 54\% of the samples, highlighting the limitations of using correlation alone to assess agreement in binary classification tasks.}
  \label{fig:mod_cons_pl_prog_left_en}
\end{figure}

Because the classifications are binary ($0$ = non-offensive, $1$ = offensive), correlation must be interpreted carefully. Low correlations don’t necessarily indicate low agreement. Binary variables can have near-zero correlation while showing substantial agreement in many cases. This is illustrated in the DeepSeek-R1 correlation matrix (Fig.~\ref{fig:reasoning_models}), where the correlation between 'Progressive Left EN' and 'Moderate Conservative PL' is $r = 0.27$, yet their upset plot (Fig.~\ref{fig:mod_cons_pl_prog_left_en}) shows agreement on 135 out of 252 samples (54\%).

\begin{figure*}
  \centering
  \begin{subfigure}[b]{\columnwidth}
    \includegraphics[height=5cm]{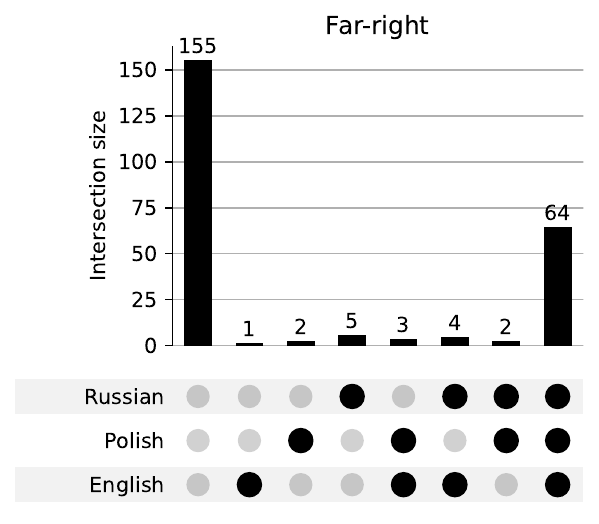}
    \caption{(a) Far-right group.}
  \end{subfigure} \hfill
  \begin{subfigure}[b]{\columnwidth}
    \includegraphics[height=5cm]{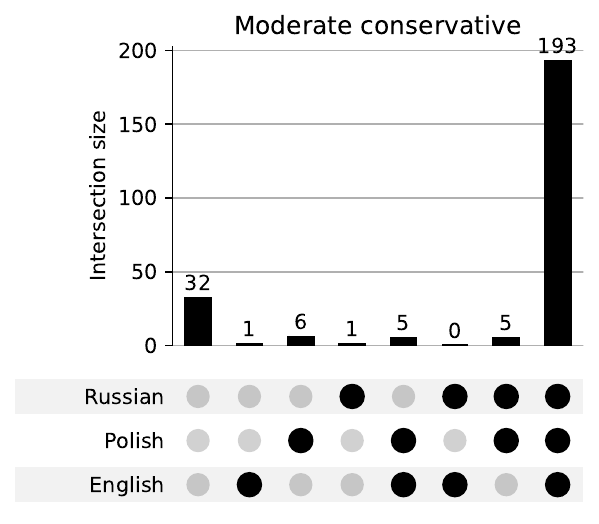}
    \caption{(b) Moderate conservative group.}
  \end{subfigure}
  
  % \vspace{1em}
  
  \begin{subfigure}[b]{\columnwidth}
    \includegraphics[height=5cm]{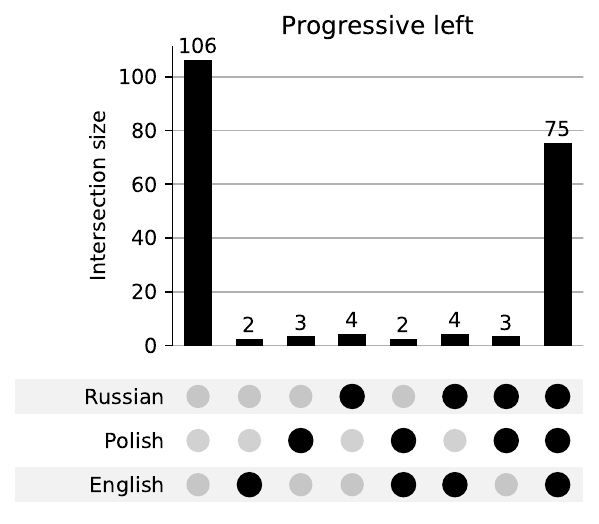}
    \caption{(c) Progressive left group.}
  \end{subfigure} \hfill
  \begin{subfigure}[b]{\columnwidth}
    \includegraphics[height=5cm]{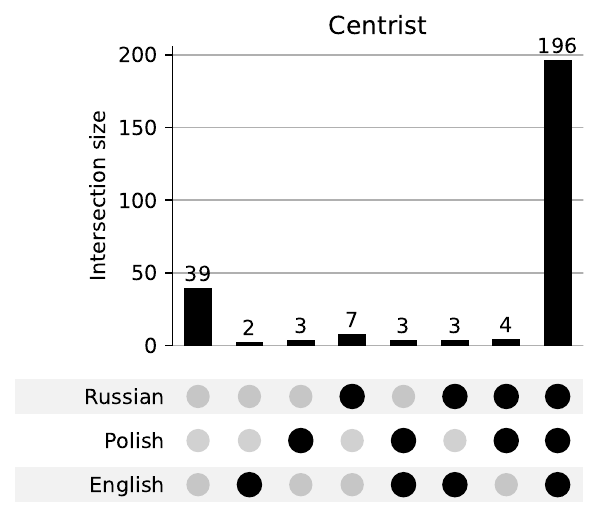}
    \caption{(d) Centrist/Independent group.}
  \end{subfigure}
  
  \caption{Upset plots showing the intersection of classification decisions across English, Polish, and Russian for each political group for the DeepSeek-R1 model's responses. These plots visualize how consistently the model classified tweets as offensive or not across different nationalities and languages within each personality profile.}
  \label{fig:upset_plots_deepseekr1}
\end{figure*}

\begin{figure*}
  \includegraphics[width=\columnwidth]{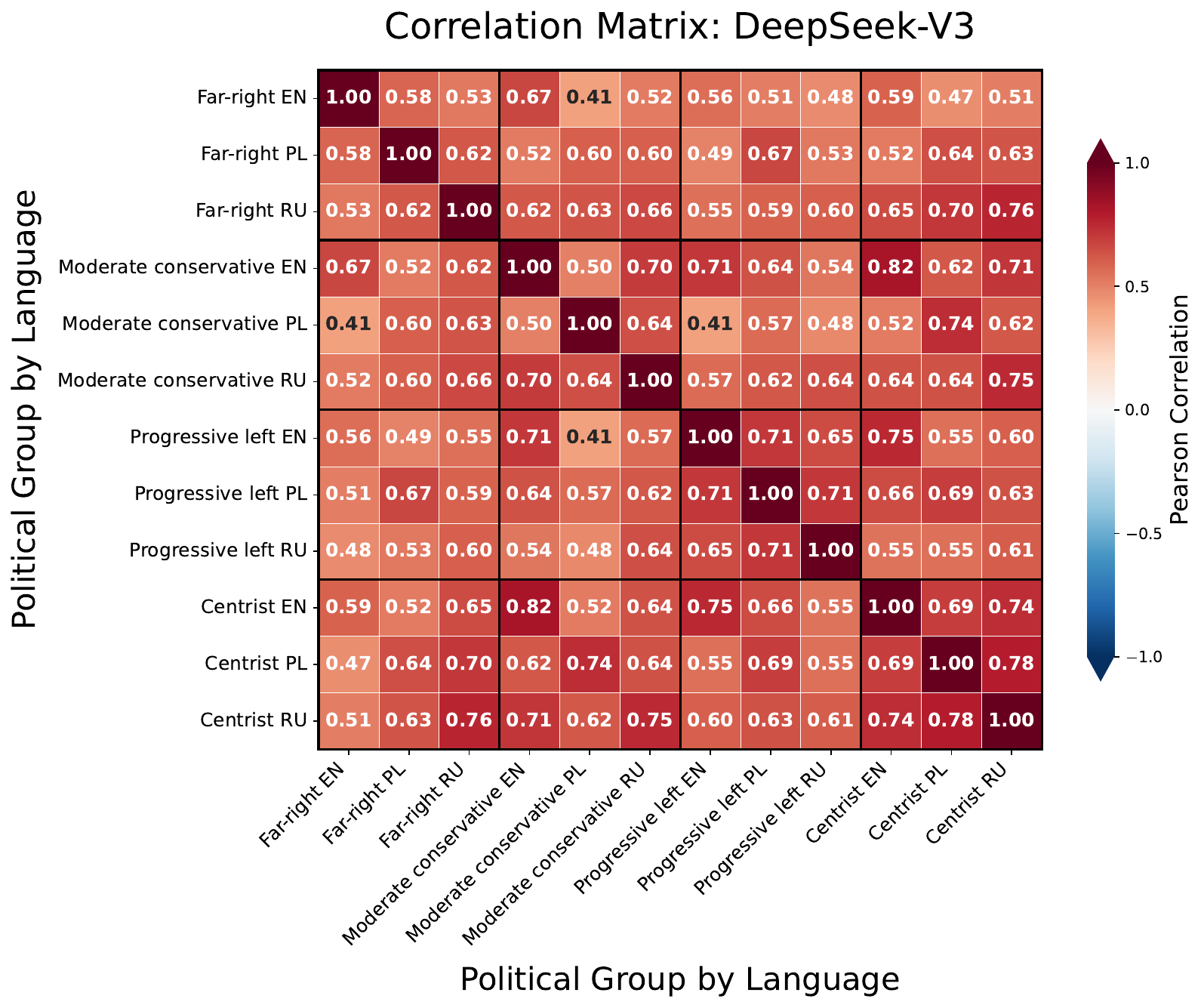} \hfill
  \includegraphics[width=\columnwidth]{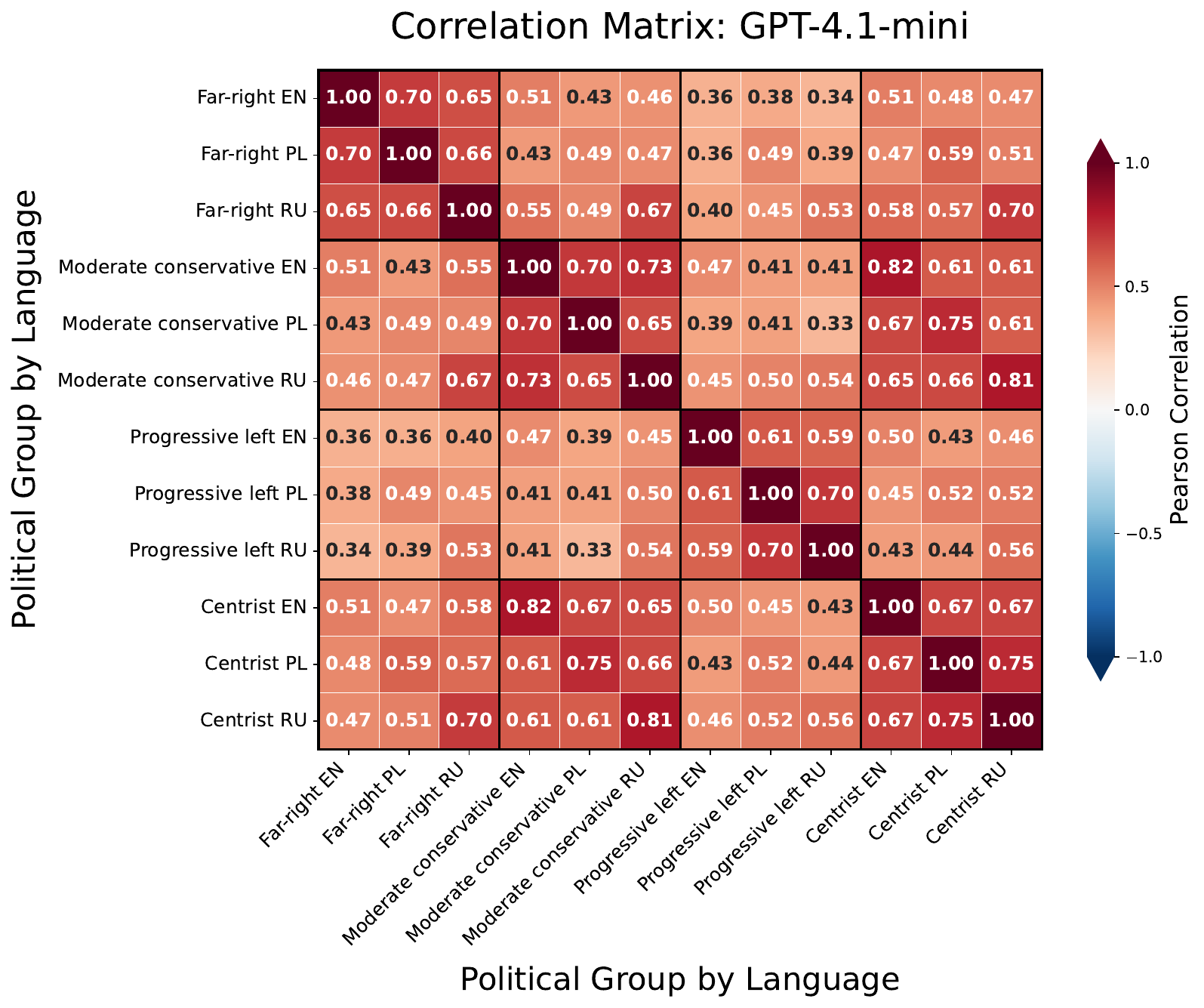}
  \caption{Correlation plots showing the agreement in tweet classifications across different political perspectives for two reasoning-disabled language models: DeepSeek-V3 and OpenAI's GPT-4.1-mini. Each plot visualizes how consistently each model assigns offensive/non-offensive labels based on the provided personality profiles, highlighting alignment or divergence across political viewpoints.}
  \label{fig:bigger_nonreasoning_models}
\end{figure*}

The upset plot (Fig.~\ref{fig:upset_plots_deepseekr1}) shows that disagreement across languages within each political group is minimal for the DeepSeek-R1 model, affecting only about $8\text{--}9\%$ of samples. Similar patterns are observed for OpenAI’s o4-mini, suggesting strong cross-language consistency and minimal influence of simulated nationality on classification.

A key strength of DeepSeek-R1 is its generation of detailed reasoning traces, while o4-mini typically provides brief, often absent, summary-like responses. To mitigate randomness, five outputs were collected per model. DeepSeek-R1 frequently explores multiple hypothetical scenarios to justify its decisions, enhancing interpretability. In contrast, o4-mini relies on templated conclusions (e.g., “The person... would likely label this as...”), offering limited insight. Notably, despite prompt language being evenly distributed (English, Polish, Russian), the language of the reasoning responses varied:

\begin{itemize}
    \item \textbf{DeepSeek-R1:} Reasoning was in English for 86\% of prompts and in Russian for 14\%. No responses were given in Polish; prompts in Polish still received reasoning in English.
    \item \textbf{OpenAI's o4-mini:} Among the 253 reasoning summaries retrieved out of 17,820 total responses, 100\% were written in English.
\end{itemize}

These findings suggest that even when models are prompted in a given language, their internal reasoning generation may default to a more dominant or better-supported language, typically English.

To quantitatively compare the two correlation matrices shown in Fig.~\ref{fig:reasoning_models}, we introduce two heuristics: one for cross-language consistency and one for inter-group differentiation. These two metrics are defined as follows:

Let \( C_{ij} = \operatorname{CorrelationSquare}(g_i, g_j) \).

\begin{align*}
    \text{CLC} &= 
    \frac{
        \sum_{i=1}^{4} \sum_{j=i}^{4} \operatorname{Var}(C_{ij})
    }{
        10^{-3} \sum_{i=1}^{4} \sum_{j=i}^{4} 1
    } \\[1em]
     \text{IGD} &= 
    \frac{
        \operatorname{Var}\left(
            \left\{ \operatorname{Mean}(C_{ij}) \;\middle|\; 1 \le i < j \le 4 \right\}
        \right)
    }{
        10^{-3}
    }
\end{align*}

\begin{itemize}
    \item \textbf{CLC} (Cross-Language Consistency) measures the variability of correlations within and between political groups across different languages.
    \item \textbf{IGD} (Inter-Group Differentiation) measures how distinct the model's responses are between different political groups, based on the average correlation values.
    \item $g_i$ refers to a political group from the set \{\textit{far-right, moderate conservative, progressive left, centrist}\}.
    \item $C_{ij}$ is the $3 \times 3$ block of correlation values representing the correlations between group $g_i$ and group $g_j$ across the three languages.
    \item \( \operatorname{Var}(C_{ij}) \) is the population variance (i.e., without Bessel's correction) of the 9 scalar entries in the matrix \( C_{ij} \), i.e.,
    \[
        \operatorname{Var}(C_{ij}) = \frac{1}{9} \sum_{a=1}^{3} \sum_{b=1}^{3} \left( C_{ij}^{(a,b)} - \mu_{ij} \right)^2
    \]
    where \( \mu_{ij} \) is the mean of all 9 entries of \( C_{ij} \).
    \item \( \operatorname{Mean}(C_{ij}) \) is the arithmetic mean of the 9 values in \( C_{ij} \).
    \item \( \operatorname{Var}(\{\cdot\}) \) denotes the population variance over the set of scalar values inside.
\end{itemize}

Lower \textbf{CLC} values indicate greater cross-language consistency, while higher \textbf{IGD} values reflect better ideological separation -- i.e., more distinct correlation patterns between political groups. \textbf{CLC} includes both intra- and inter-group comparisons, whereas \textbf{IGD} considers only inter-group pairs.

\begin{table*}[hbt!]
\centering
\Large
\resizebox{\textwidth}{!}{%
\begin{tabular}{|l|l|l|l|l|l|l|l|l|}
\hline
 & DeepSeek-R1 & OpenAI's o4-mini & DeepSeek-V3 & Qwen3-8B & OpenAI's GPT-4.1-mini & Qwen3-4B & Gemma3-4B-IT & Mistral-7B-Instruct-v0.3 \\ \hline
Category & \begin{tabular}[c]{@{}l@{}}Big\\ reasoning\end{tabular} & \begin{tabular}[c]{@{}l@{}}Big\\ reasoning\end{tabular} & \begin{tabular}[c]{@{}l@{}}Big\\ non-reasoning\end{tabular} & \begin{tabular}[c]{@{}l@{}}Small\\ reasoning\end{tabular} & \begin{tabular}[c]{@{}l@{}}Small\\ non-reasoning\end{tabular} & \begin{tabular}[c]{@{}l@{}}Small\\ non-reasoning\end{tabular} & \begin{tabular}[c]{@{}l@{}}Small\\ non-reasoning\end{tabular} & \begin{tabular}[c]{@{}l@{}}Small\\ non-reasoning\end{tabular} \\ \hline
\begin{tabular}[c]{@{}l@{}}Percentage of \\ valid responses (\%)\end{tabular} & 90.7 & 90.7 & 100 & 77.4 & 100 & 100 & 100 & 100 \\ \hline
\begin{tabular}[c]{@{}l@{}}Cross-Language  \\ Consistency (CLC) \\ (lower is better)\end{tabular} & \textbf{3.92} & 4.85 & 15.31 & 22.20 & 12.32 & 33.43 & 28.29 & 65.64\\ \hline
\begin{tabular}[c]{@{}l@{}}Inter-Group \\ Differentiation (IGD) \\ (higher is better)\end{tabular} & \textbf{100.03} & 89.28 & 1.58 & 32.23 & 8.09 & 4.77 & 3.46 & 1.37\\ \hline
\end{tabular}
}
\caption{Comparison of models}
\label{tab:model_comparison}
\end{table*}

In our results (Table~\ref{tab:model_comparison}), DeepSeek-R1 achieves a \textbf{CLC} of $3.92$ and \textbf{IGD} of $100.03$, compared to o4-mini’s $4.85$ and $89.28$, respectively. This indicates that DeepSeek-R1 is more consistent across languages and better at distinguishing between political ideologies.

Overall, both models show strong consistency (Fig.~\ref{fig:reasoning_models}). They capture political context effectively, producing structured, coherent, and personalized classification outputs, with DeepSeek-R1 showing a slight edge in clarity and stability.

\subsection{Large Models without Reasoning Capabilities}

Despite its size, the non-reasoning model DeepSeek-V3 produces largely uninformative correlation plots (Fig.~\ref{fig:bigger_nonreasoning_models}), with all values positive and high (lowest $r = 0.41$). As discussed in Section~\ref{bigger_reasoning_models}, high correlation can sometimes mask important differences; here, it reflects broad agreement across outputs rather than meaningful differentiation.

A key concern is that intra-group correlations (same ideology across languages) are nearly identical to inter-group correlations (different ideologies), suggesting minimal ideological differentiation. For instance, far-right responses correlate almost equally with progressive left outputs, highlighting DeepSeek-V3's limited ability to personalize. 

This is further confirmed by the \textbf{CLC} and \textbf{IGD} scores: DeepSeek-V3 yields $15.31$ (CLC) and $1.58$ (IGD) (Table~\ref{tab:model_comparison}). Overall, DeepSeek-V3 underperforms compared to reasoning-enabled models, showing high consistency across outputs but poor discrimination between ideological groups. 

\subsection{Smaller Model with Reasoning Capabilities}

Integrated reasoning in LLMs is a recent innovation, primarily appearing in large-scale models. Few small models currently support explicit reasoning. In our study, we evaluated Qwen3-8B -- one of the newest small models with configurable reasoning, which we enabled for all experiments.

Qwen3-8B demonstrates reasoning ability in the prompt language, though with a strong English bias. Across evenly distributed prompts (English, Polish, Russian), reasoning outputs were in English for 62.8\% of samples, Russian for 33.0\%, and Polish for just 4.2\%.

\begin{figure}
  \includegraphics[width=\columnwidth]{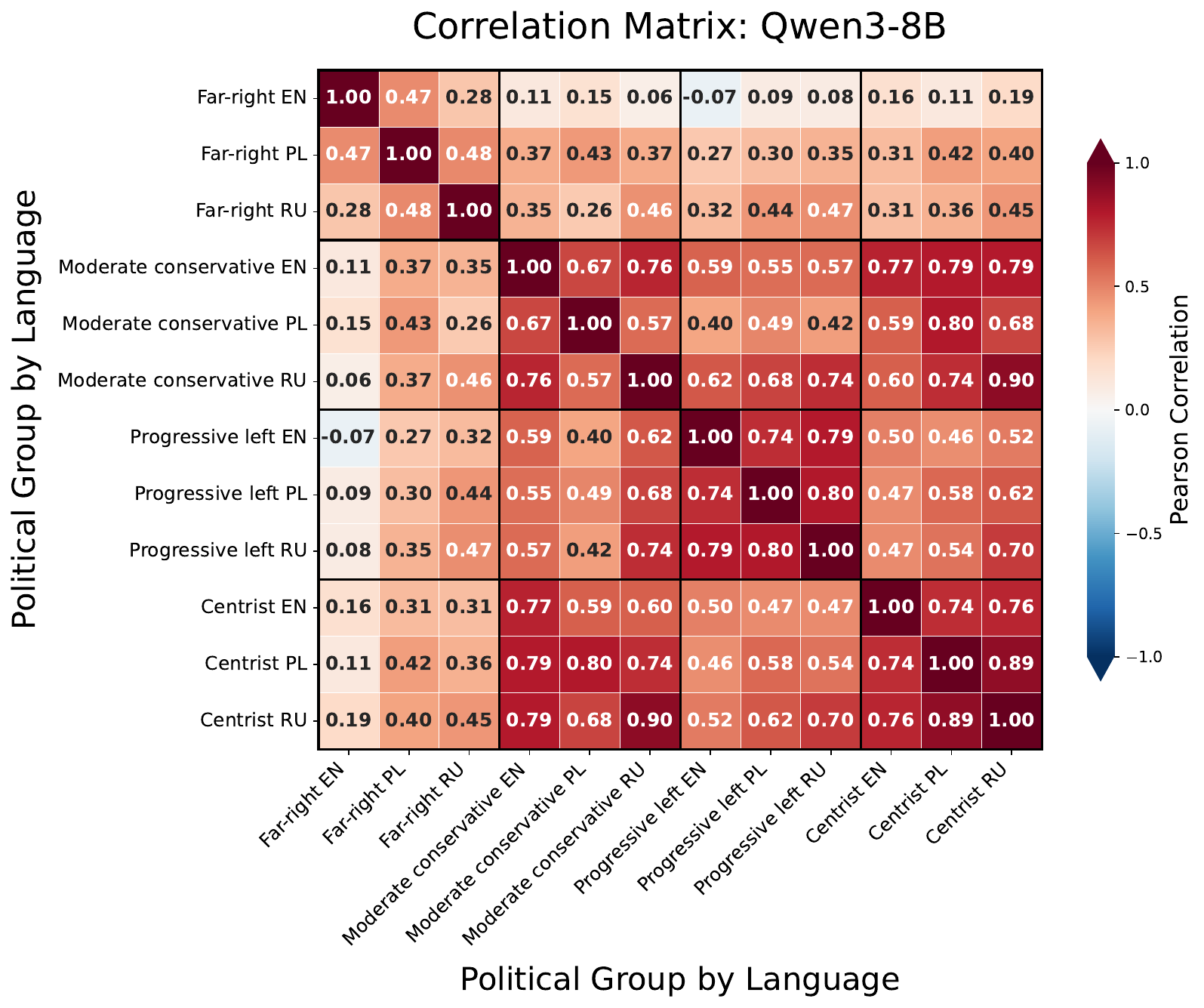}
  \caption{Correlation plots showing the agreement in tweet classifications across different political perspectives for reasoning-enabled Qwen3-8B model.}
  \label{fig:qwen_reasoning}
\end{figure}

The model’s correlation plot (Fig.~\ref{fig:qwen_reasoning}) shows mostly positive values, with one negative outlier. The strongest between-group similarity is between Moderate Conservative and Centrist groups, while the Far-Right group shows the lowest average correlation with others -- indicating greater differentiation in its personalized classifications.

\subsection{Smaller Models without Reasoning Capabilities}

We evaluated several non-reasoning models, ranging from small LLMs to larger variants, including Gemma 3-4B-IT (Google), Qwen3-4B (Alibaba), Mistral-7B-Instruct-v0.3~\cite{mistral7b, mistral7b_instruct_v03}, and GPT-4.1-mini~\cite{openai2025gpt41}.

\begin{figure}
  % \centering
  \includegraphics[width=0.9\columnwidth]{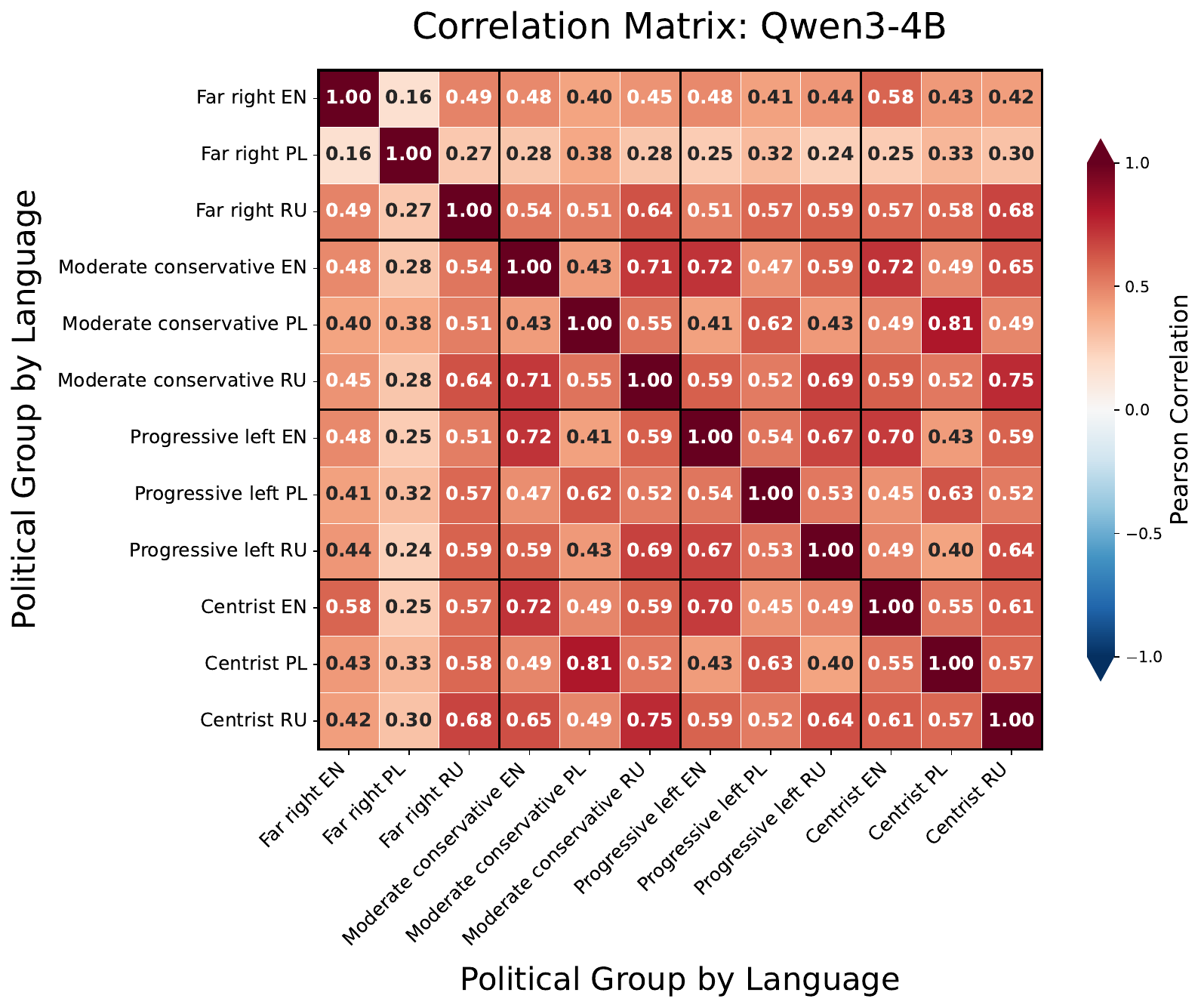}
  
  \includegraphics[width=0.9\columnwidth]{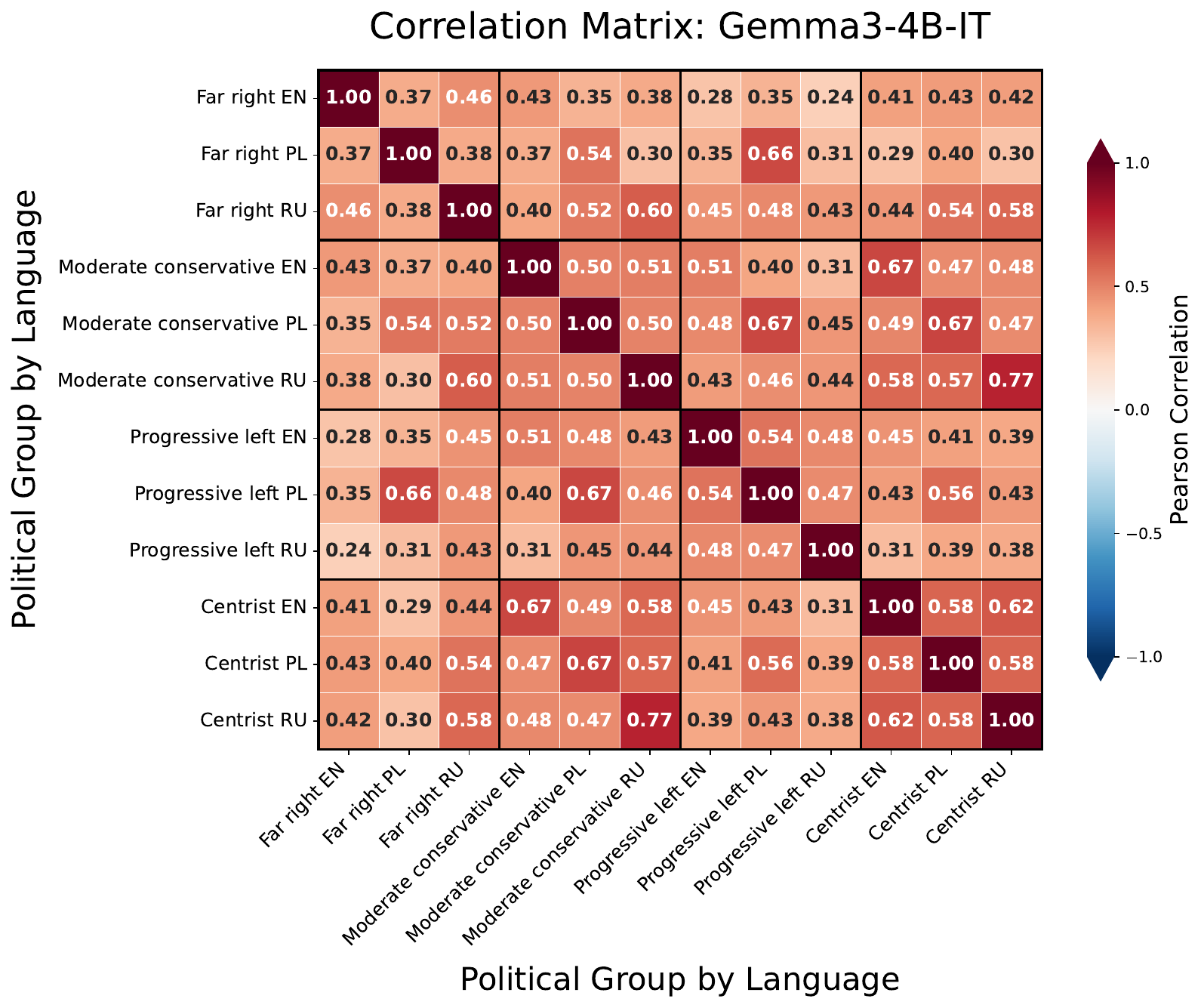}
  
  \includegraphics[width=0.9\columnwidth]{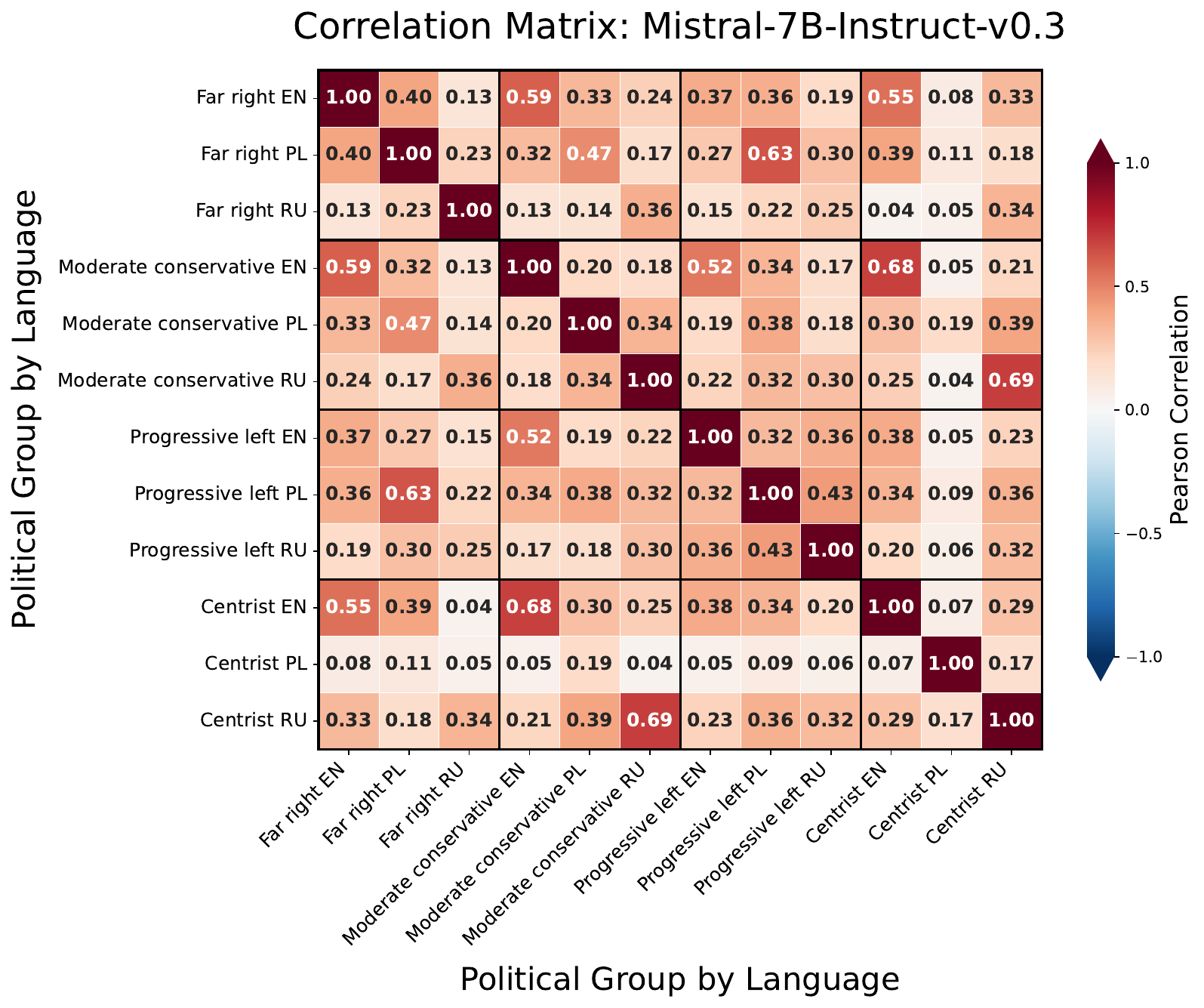}
  
  \caption{Correlation matrices for small non-reasoning models: Qwen3-4B, Gemma3-4B-IT, and Mistral-7B-Instruct-v0.3. All models exhibit generally high positive correlations, suggesting limited differentiation between political viewpoints.}
  \label{fig:smaller_models_without_reasoning}
\end{figure}

As shown in Fig.~\ref{fig:smaller_models_without_reasoning}, all small models exhibit uniformly positive correlations, indicating alignment across prompts regardless of language, personality, or nationality. This suggests limited ability to distinguish nuanced political perspectives.

Gemma3-4B-IT and Qwen3-4B show consistently high correlations, whereas Mistral-7B-Instruct-v0.3 displays more variability ($0.04 \leq r \leq 0.69$), indicating less stable behavior. Notably, all models show a high correlation between 'Moderate Conservative Russian' and 'Centrist Russian' responses, which is likely due to random variation rather than meaningful ideological similarity.

Extending the analysis to GPT-4.1-mini, it exhibits broadly positive correlations similar to the models above but with slightly more variability, suggesting modest improvements in distinguishing outputs across ideological groups. Compared to DeepSeek-V3, which shows minimal differentiation, GPT-4.1-mini achieves better group separation while maintaining relatively high consistency.

This trend is also reflected in the \textbf{CLC} and \textbf{IGD} scores:  models like Gemma3-4B-IT, Qwen3-4B, and Mistral-7B-Instruct-v0.3 have modest scores, whereas GPT-4.1-mini achieves $12.32$ (CLC) and $8.09$ (IGD) (Table~\ref{tab:model_comparison}), indicating improved but still limited ideological discrimination. Overall, small non-reasoning models fall short of reasoning-enabled models in capturing nuanced political distinctions.

\section{Discussion}

Our results show reasoning capabilities are crucial for large language models (LLMs) to simulate ideologically and culturally grounded interpretations of offensive content. Models like DeepSeek-R1 and o4-mini aligned more closely with ideological framing in prompts, generating outputs that varied meaningfully across far-right, centrist, and progressive personas. Correlation analyses confirmed strong within-group consistency and clear ideological separation, with this effect persisting across English, Polish, and Russian prompts, highlighting the role of reasoning in cross-linguistic coherence.

DeepSeek-R1 stood out for detailed, coherent output, offering interpretable insights into its decision-making. In contrast, o4-mini produced more generic, templated explanations, indicating reasoning depth and variability—not just presence—are essential for simulating personalized language understanding.

Non-reasoning models like DeepSeek-V3 and GPT-4.1-mini showed little sensitivity to ideological framing, with outputs mostly uniform across personas. This suggests that model scale alone doesn't yield sociopolitical nuance. Even smaller reasoning-enabled models struggled with ideological differentiation, though Qwen3-8B showed modest gains.

A notable pattern was the dominance of English in reasoning outputs, even when prompts were in Polish or Russian. This reflects a multilingual alignment gap that may hinder interpretability for non-English users.

The primary risk of this work lies in reinforcing ideological biases through simulated personas, which may inadvertently legitimize harmful viewpoints if not carefully framed. Additionally, reliance on LLM-generated reasoning in sensitive political contexts could amplify existing stereotypes or overlook nuanced cultural interpretations, especially given the models' tendency to default to English reasoning.

Finally, we found that Pearson correlations, while helpful, can be misleading in binary tasks. High agreement can coincide with low correlation, reinforcing the value of complementary tools like upset plots for analyzing model behavior.

\section{Conclusions and Future Work}

This study examined how large language models can simulate personalized offensiveness judgments based on ideological and cultural cues. We proposed a novel framework combining persona-based prompting, multilingual analysis, and reasoning-driven classification. Results highlight reasoning as a key factor for producing ideologically sensitive and interpretable outputs. Models like DeepSeek-R1 not only captured political perspectives effectively but also showed consistent behavior across languages and nationalities.

These findings underscore the importance of reasoning for nuanced, human-like judgments, especially in politically or culturally sensitive contexts. Reasoning-enabled models provide more transparent justifications, which are vital for trustworthy NLP applications.

Future work should expand the dataset to cover more sociopolitical viewpoints and languages, and consider real user data or crowd-annotated profiles for realism. Exploring few-shot and retrieval-augmented techniques could further personalize responses. There's also a need for better benchmarks to evaluate reasoning quality, especially in multilingual settings where disparities remain challenging.

This study was also conducted as a pilot within the InteGra system, focusing on methods for defining different personas and analyzing decision-making differences between them in identical contexts. Future work will extend into game studies, incorporating additional psychological aspects of characters as well as their roles in dialogue contexts.

\section*{Limitations}

While our findings provide compelling evidence for the value of reasoning in personalized offensiveness detection, several limitations must be considered. First, the dataset used in this study consisted of a relatively small sample -- 300 tweets from a single political domain. This limited scope may constrain the generalizability of our results and does not capture the full spectrum of offensive content.

Second, the personas were manually constructed and necessarily idealized. Although this allowed for controlled experimentation, it does not fully reflect the diversity, inconsistency, and contextual variation of real-world individuals. Future work should aim to incorporate more empirically grounded or dynamically generated persona representations.

Third, while translations were carefully reviewed, a small number of errors and edge cases may have influenced the multilingual evaluation. Cultural subtleties are difficult to preserve in translation, and it remains possible that some tweets were interpreted differently due to linguistic artifacts rather than genuine ideological reasoning.

We also acknowledge limitations imposed by the models themselves. For instance, models like o4-mini are proprietary and offer limited control over generation parameters such as temperature, affecting reproducibility and interpretability. Furthermore, we observed a consistent bias toward generating reasoning in English, regardless of input language. This reduces authenticity of multilingual outputs and hinders inclusive evaluation across diverse language communities.

Lastly, the binary classification framing used in this study simplifies offensiveness into a yes/no decision, ignoring gradations of harm, context, or intent. While useful for structured analysis, this dichotomy overlooks the complexity inherent in real-world moderation and annotation. Addressing these issues in future work will be essential for building more comprehensive and equitable personalization frameworks.

\section*{Acknowledgments}
%CLARIN-PL % TODO AFTER REVIEW
This work was financed by
(1) the European Regional Development Fund as part of the 2021–2027 European Funds for a Modern Economy (FENG) programme, project no. FENG.02.04-IP.040004/24: CLARIN – Common Language Resources and Technology Infrastructure;
(2) CLARIN-PL (POIR.04.02.00-00C002/19);
(3) DARIAH-PL (POIR.04.02.00-00-D006/20, KPOD.01.18-IW.03-0013/23);
(4) CLARIN ERIC – European Research Infrastructure Consortium: Common Language Resources and Technology Infrastructure (period: 2024-2026) funded by the Polish Minister of Science under the programme: "Support for the participation of Polish scientific teams in international research infrastructure projects", agreement number 2024/WK/01;
(5) Polish Minister of Digital Affairs under a special purpose subsidy No. 1/WII/DBI/2025: HIVE AI: Development and Pilot Deployment of Large Language Models in the Polish Public Administration;
(6) the statutory funds of the Department of Artificial Intelligence, Wroclaw University of Science and Technology;
(7) the National Science Centre, Poland, project no. 2021/41/B/ST6/04471;
(8) project no. FENG.01.01-IP.02-2061/23, co-financed by the European Union under the European Funds for Smart Economy 2021–2027 programme.

AI-based tools, including ChatGPT, Grammarly Premium, and Writeful, were used exclusively to support linguistic clarity and improve the readability of the manuscript.

\bibliographystyle{IEEEtran}
\bibliography{IEEEfull, custom}

\appendix

\section{Additional Study of Distributions}

Additionally, we extracted the probabilities of generating the response $1$ (offensive) and $0$ (non-offensive) from the smaller models without reasoning capabilities (Gemma3-4B-IT, Qwen3-4B, Mistral-7B-Instruct-v0.3). While the final output of a language model is technically a multinomial distribution over all tokens, we observed that in only $0.4\%$ of cases (5 samples), the combined probability of generating either $0$ or $1$ differed from $1.0$ by more than $0.01$. This suggests that the assumption of a near-binomial distribution between the two target classes is accurate for the purpose of this analysis.

We also plotted the distributions of probabilities for generating the response $1$ (offensive) from each of the three models, across all 12 different prompt configurations (4 political personas in 3 languages) (see the plots at \href{https://github.com/DzmitryPihulski/2025_SENTIRE_data/blob/main/data/models_results/Gemma3_4b/first/temp_1/distributions.pdf}{Gemma3\_4b plot}, \href{https://github.com/DzmitryPihulski/2025_SENTIRE_data/blob/main/data/models_results/Mistral_7b/first/temp_1/distributions.pdf}{Mistral\_7b plot}, \href{https://github.com/DzmitryPihulski/2025_SENTIRE_data/blob/main/data/models_results/Qwen3_4b/first/temp_1/distributions.pdf}{Qwen3\_4b plot} in our GitHub repository~\cite{repo2025}). The analysis of these plots suggests that the models are generally very confident in their predictions. Specifically, in 84\% of the evaluated cases, the predicted probability for the offensive class was either very high ($\ge 0.95$) or very low ($\le 0.05$), indicating strong classification certainty across most inputs. 
Among the three models, Gemma3-4B-IT is the most confident on average, with 89\% of evaluated cases having predicted probabilities for the offensive class either very high ($\ge 0.95$) or very low ($\le 0.05$) (see the plot at \href{https://github.com/DzmitryPihulski/2025_SENTIRE_data/blob/main/data/models_results/Gemma3_4b/first/temp_1/distributions.pdf}{Gemma3\_4b plot} in the repository~\cite{repo2025}). However, an inspection of the probability distributions reveals a strong bias toward classifying tweets as offensive across all political groups and languages. This suggests that the model exhibits a generally low tolerance for hate speech, even when prompted to act as a persona that downplays offensiveness. Specifically, Gemma3-4B-IT produced a probability $> 0.5$ for the offensive class in 2,621 out of 3,564 samples. The Qwen3-4B model exhibits a nearly identical behavior to Gemma3-4B-IT in terms of confidence and classification patterns (see the plot at \href{https://github.com/DzmitryPihulski/2025_SENTIRE_data/blob/main/data/models_results/Qwen3_4b/first/temp_1/distributions.pdf}{Qwen3\_4b plot} in the repository~\cite{repo2025}). The only model that assigns low probabilities to the offensive class ($\le 0.05$) in a notable number of samples is Mistral-7B-Instruct-v0.3 (see the plot at \href{https://github.com/DzmitryPihulski/2025_SENTIRE_data/blob/main/data/models_results/Mistral_7b/first/temp_1/distributions.pdf}{Mistral\_7b plot} in the repository~\cite{repo2025}). In particular, for two categories -- \textit{Far-right English} and \textit{Centrist English} -- the number of samples with low offensive probabilities exceeds those with high probabilities.

\end{document}